%% file: eacl2021.tex
\documentclass[11pt,a4paper]{article}
\usepackage[hyperref]{eacl2021}
\usepackage{times}
\usepackage{latexsym}

\usepackage{microtype}

\usepackage{amsmath}
\usepackage{mathtools}
\usepackage{booktabs}
\usepackage{graphicx}
\aclfinalcopy 
\def\aclpaperid{1003} 

\title{Unsupervised Abstractive Summarization of Bengali Text Documents\\}

\author{
Radia Rayan Chowdhury\thanks{~~Equal contribution. Listed by alphabetical order.} \\
Ahsanullah Univ of Science \& Tech \\
Dhaka, Bangladesh \\
\texttt{radiarayan.rrc@gmail.com} \\
\And
Mir Tafseer Nayeem\footnotemark[1] \\
Ahsanullah Univ of Science \& Tech \\
Dhaka, Bangladesh \\
\texttt{mir.nayeem@alumni.uleth.ca} \\
\AND 
Tahsin Tasnim Mim \\
Ahsanullah Univ of Science \& Tech \\
Dhaka, Bangladesh \\
\texttt{tahsintasnimmim@gmail.com} \\
\And
Md. Saifur Rahman Chowdhury \\
Ahsanullah Univ of Science \& Tech \\
Dhaka, Bangladesh \\
\texttt{saif.chowdhury1997@gmail.com} \\
\AND 
Taufiqul Jannat \\
Ahsanullah Univ of Science \& Tech \\
Dhaka, Bangladesh \\
\texttt{taufiquljannat@gmail.com} \\
}

\date{}

\begin{document}

\maketitle

\begin{abstract}
Abstractive summarization systems generally rely on large collections of document-summary pairs. However, the performance of abstractive systems remains a challenge due to the unavailability of the parallel data for low-resource languages like Bengali. To overcome this problem, we propose a graph-based unsupervised abstractive summarization system in the single-document setting for Bengali text documents, which requires only a Part-Of-Speech (\textbf{POS}) tagger and a pre-trained language model trained on Bengali texts. We also provide a human-annotated dataset with document-summary pairs to evaluate our abstractive model and to support the comparison of future abstractive summarization systems of the Bengali Language. We conduct experiments on this dataset and compare our system with several well-established unsupervised extractive summarization systems. Our unsupervised abstractive summarization model outperforms the baselines without being exposed to any human-annotated reference summaries.\footnote{We make our code \& dataset publicly available at \url{https://github.com/tafseer-nayeem/BengaliSummarization} for reproduciblity.}
\end{abstract}

\section{Introduction}
The process of shortening a large text document with the most relevant information of the source is known as automatic text summarization. A good summary should be coherent, non-redundant, and grammatically readable while retaining the original document's most important contents \citep{nenkova2012survey, nayeem2018abstractive}. There are two types of summarizations: extractive and abstractive. Extractive summarization is about ranking important sentences from the original text. The abstractive method generates human-like sentences using natural language generation techniques. Traditionally used abstractive techniques are sentence compression, syntactic reorganization, sentence fusion, and lexical paraphrasing \cite{DBLP:conf/aaai/LinN19}. Compared to extractive, abstractive summary generation is indeed a challenging task.

A cluster of sentences uses multi-sentence compression (\textbf{MSC}) to summarize into one single sentence originally called sentence fusion \citep{barzilay2005sentence, 10.1145/3132847.3133106}. The success of neural sequence-to-sequence (\textbf{seq2seq}) models with attention \cite{bahdanau2014neural, luong-pham-manning:2015:EMNLP} provides an effective way for text generation which has been extensively applied in the case of abstractive summarization of English language documents \cite{rush2015neural, chopra-etal-2016-abstractive, nallapati-etal-2016-abstractive, miao-blunsom-2016-language, paulus2017deep, DBLP:conf/ecir/NayeemFC19}. These models are usually trained with lots of gold summaries, but there is no large-scale human-annotated abstractive summaries available for low-resource language like Bengali. In contrast, the unsupervised approach reduces the human effort and cost for collecting and annotating large amount of paired training data. Therefore, we choose to create an effective Bengali Text Summarizer with an unsupervised approach. The summary of our contributions:

\begin{itemize}
    \item To the best of our knowledge, our \textbf{Ben}gali Text \textbf{Summ}arization model (\textbf{BenSumm}) is the  very first unsupervised model to generate abstractive summary from Bengali text documents while being simple yet robust.
    
    \item We also introduce a  highly abstractive dataset with document-summary pairs to evaluate our model, which is written by professional summary writers of National Curriculum and Textbook Board (\textbf{NCTB}).\footnote{\url{http://www.nctb.gov.bd/}}
    
    \item We design an unsupervised abstractive sentence generation model that performs sentence fusion on Bengali texts. Our model requires only \textbf{POS} tagger and a pre-trained language model, which is easily reproducible.

\end{itemize}

\section{Related works}

Many researchers have worked on text summarization and introduced different extractive and abstractive methods. Nevertheless, very few attempts have been made for Bengali Text summarization despite Bangla being the $7^{th}$ most spoken language.\footnote{\url{https://w.wiki/57}} \citet{das2010topic} developed Bengali opinion based text summarizer using given topic which can determine the information on sentiments of the original texts. \citet{haque2017innovative, haque2015automatic} worked on extractive Bengali text summarization using pronoun replacement, sentence ranking with term frequency, numerical figures, and overlapping of title words with the document sentences. Unfortunately, the methods are limited to extractive summarization, which ranks some important sentences from the document instead of generating new sentences which is challenging for an extremely low resource language like Bengali. Moreover, there is no human-annotated dataset to compare abstractive summarization methods of this language.

\citet{jing2000cut} worked on Sentence Compression (\textbf{SC}) which has received considerable attention in the NLP community. Potential utility for extractive text summarization made \textbf{SC} very popular for single or multi-document summarization \citep{nenkova2012survey}. TextRank \citep{mihalcea2004textrank} and LexRank \citep{erkan2004lexrank} are graph-based methods for extracting important sentences from a document. \citet{clarke2008global, filippova2010multi} showed a first intermediate step towards abstractive summarization, which compresses original sentences for a summary generation. The Word-Graph based approaches were first proposed by \cite{filippova2010multi}, which require only a POS tagger and a list of stopwords. \citet{boudin2013keyphrase} improved Filippova’s approach by re-ranking the compression paths according to keyphrases, which resulted in more informative sentences. \citet{nayeem2018abstractive} developed an unsupervised abstractive summarization system that jointly performs sentence fusion and paraphrasing.

\section{BenSumm Model}

We here describe each of the steps involved in our \textbf{Ben}gali Unsupervised Abstractive Text \textbf{Summ}arization model (\textbf{BenSumm}) for single document setting. Our preprocessing step includes tokenization, removal of stopwords, Part-Of-Speech (\textbf{POS}) tagging, and filtering of punctuation marks. We use the \textbf{NLTK}\footnote{\url{https://www.nltk.org}} and \textbf{BNLP}\footnote{\url{https://bnlp.readthedocs.io/en/latest/}} to preprocess each sentence and obtain a more accurate representation of the information.

\subsection{Sentence Clustering}
The clustering step allows us to group similar sentences from a given document. This step is critical to ensure good coverage of the whole document and avoid redundancy by selecting at most one sentence from each cluster \cite{nayeem2017extract}.  The Term Frequency-Inverse Document Frequency (\textbf{TF-IDF}) measure does not work well \cite{aggarwal2012survey}. Therefore, we calculate the cosine similarity between the sentence vectors obtained from \textbf{ULMfit} pre-trained language model \citep{howard2018universal}. We use hierarchical agglomerative clustering with the ward’s method \cite{murtagh2014ward}. There will be a minimum of 2 and a maximum of $n-1$ clusters. Here, $n$ denotes the number of sentences in the document. We measure the number of clusters for a given document using the silhouette value. The clusters are highly coherent as it has to contain sentences similar to every other sentence in the same cluster even if the clusters are small. The following formula can measure silhouette Score:

\begin{equation}
\text{Silhouette Score} = \frac{(x-y)}{max(x,y)}
\end{equation}

where $y$ denotes mean distance to the other instances of intra-cluster and $x$ is the mean distance to the instances of the next closest cluster.

\subsection{Word Graph (WG) Construction}

\begin{figure}[htbp]
    \centering
    \includegraphics[scale = 1.1]{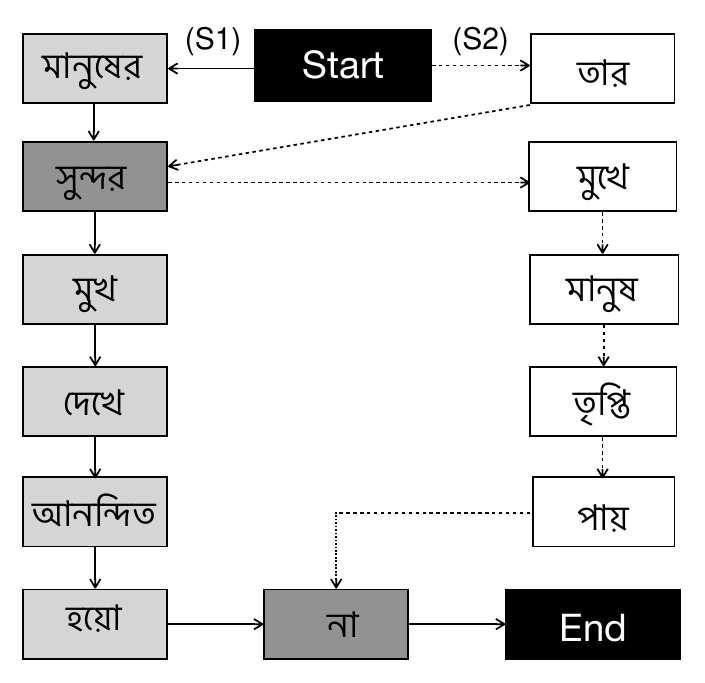}
    \caption{Sample WG of two related sentences.}
    \label{fig:word_graph}
\end{figure}

Textual graphs to generate abstractive summaries provide effective results \citep{ganesan2010opinosis}. We chose to build an abstractive summarizer with a sentence fusion technique by generating word graphs \cite{filippova2010multi, boudin2013keyphrase} for the Bengali Language. This method is entirely unsupervised and needs only a \textbf{POS} tagger, which is highly suitable for the low-resource setting. Given a cluster of related sentences, we construct a word-graph following \cite{filippova2010multi,boudin2013keyphrase}. Let, a set of related sentences S = \{$s_1$, $s_2$, ..., $s_n$\}, we construct a graph $G = (V, E)$ by iteratively adding sentences to it. The words are represented as vertices along with the parts-of-speech (\textbf{POS}) tags. Directed edges are formed by connecting the adjacent words from the sentences. After the first sentence is added to the graph as word nodes (punctuation included), words from the other related sentences are mapped onto a node in the graph with the same \textbf{POS} tag. Each sentence of the cluster is connected to a dummy start and end node to mark the beginning and ending sentences. After constructing the word-graph, we can generate $M$-shortest paths from the dummy start node to the end node in the word graph (see Figure \ref{fig:word_graph}).

\begin{figure}[htbp]
    \centering
    \includegraphics[scale = 0.48]{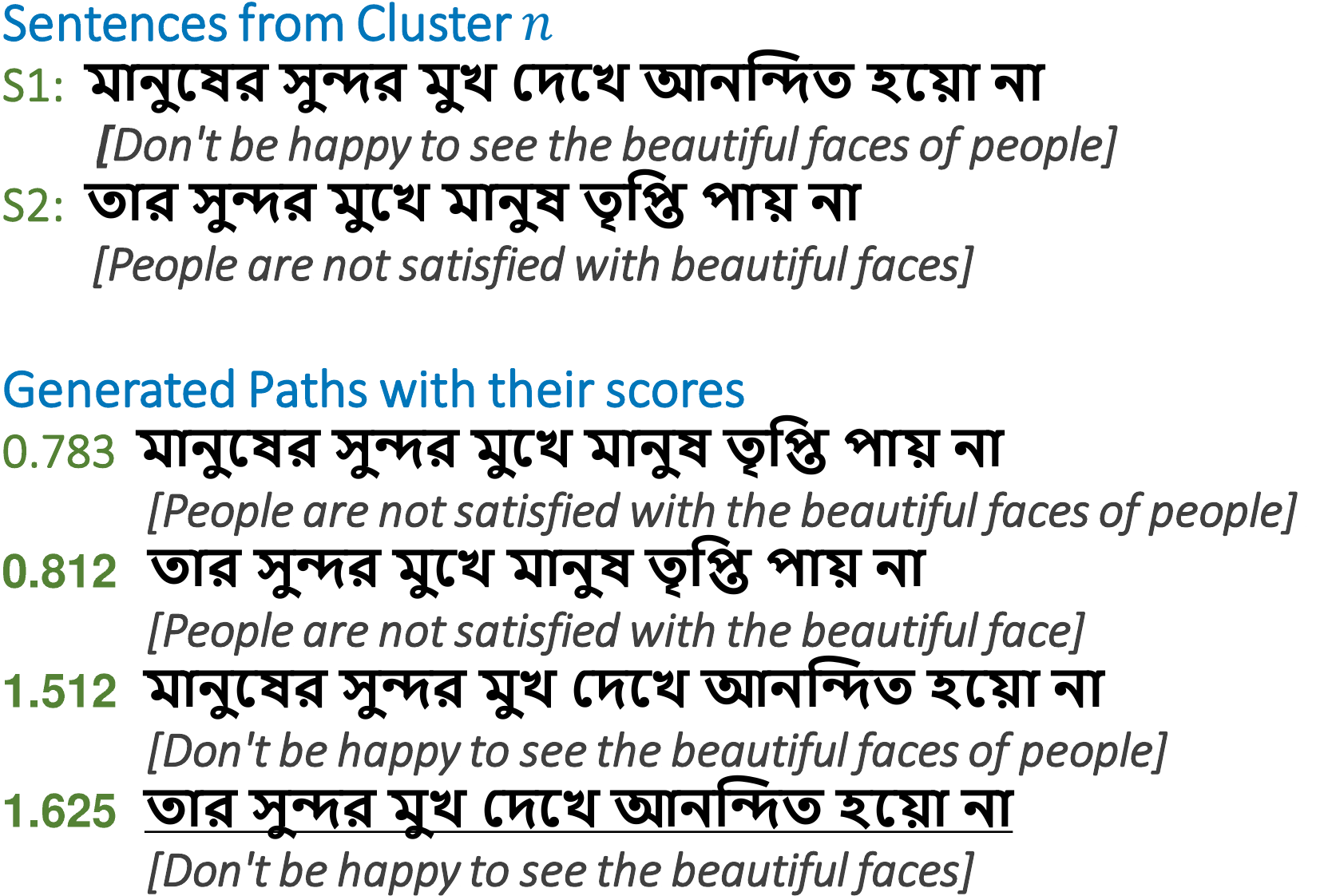}
    \caption{Output of WG given two related sentences. The underlined sentence is the top-ranked sentence to be included in the final summary.}
    \label{fig:WG_example}
\end{figure}

Figure \ref{fig:WG_example} presents two sentences, which is one of the source document clusters, and the possible paths with their weighted values are generated using the word-graph approach. Figure \ref{fig:word_graph} illustrates an example \textbf{WG} for these two sentences.

\begin{figure}[htbp]
    \centering
    \includegraphics[scale = 0.80]{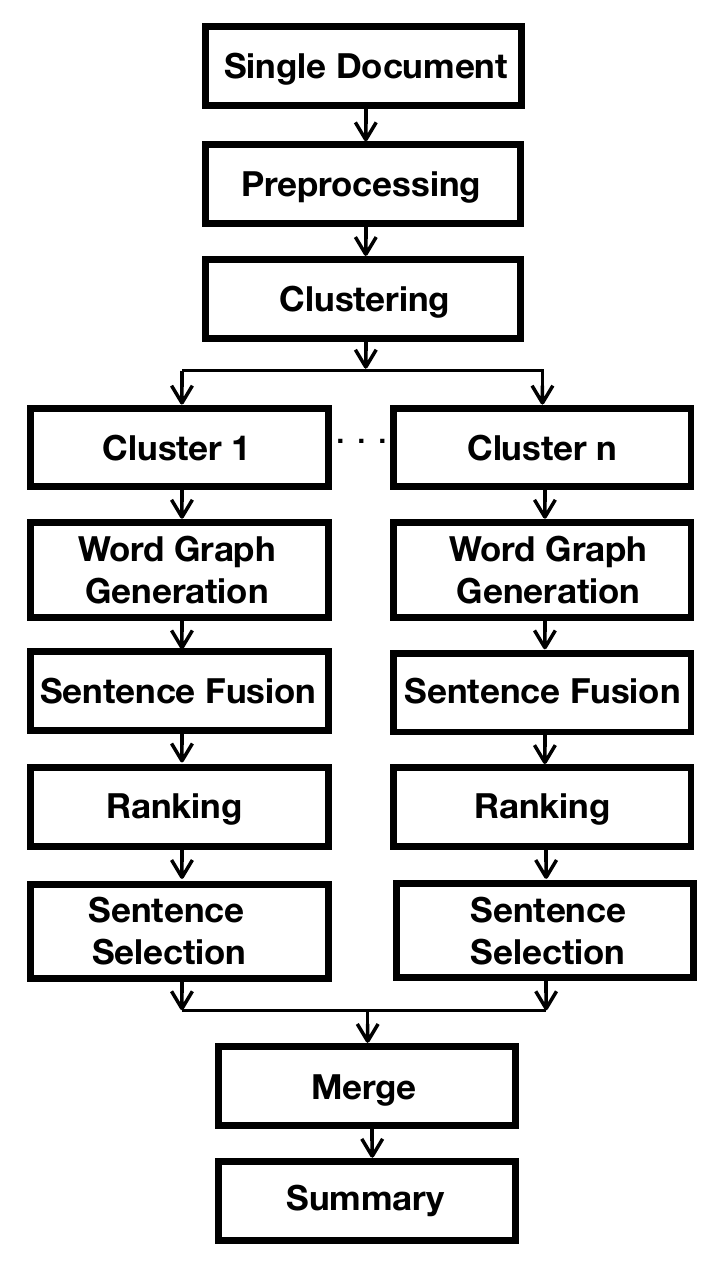}
    \caption{Overview of our \textbf{BenSumm} model.}
    \label{fig:BenSumm_model}
\end{figure}

After constructing clusters given a document, a word-graph is created for each cluster to get abstractive fusions from these related sentences. We get multiple weighted sentences (see Figure \ref{fig:WG_example}) form the clusters using the ranking strategy \cite{boudin2013keyphrase}. We take the top-ranked sentence from each cluster to present the summary. We generate the final summary by merging all the top-ranked sentences. The overall process is presented in Figure \ref{fig:BenSumm_model}. We also present a detailed illustration of our framework with an example source document in the Appendix.

\begin{figure}[htbp]
    \centering
    \includegraphics[scale = 0.48]{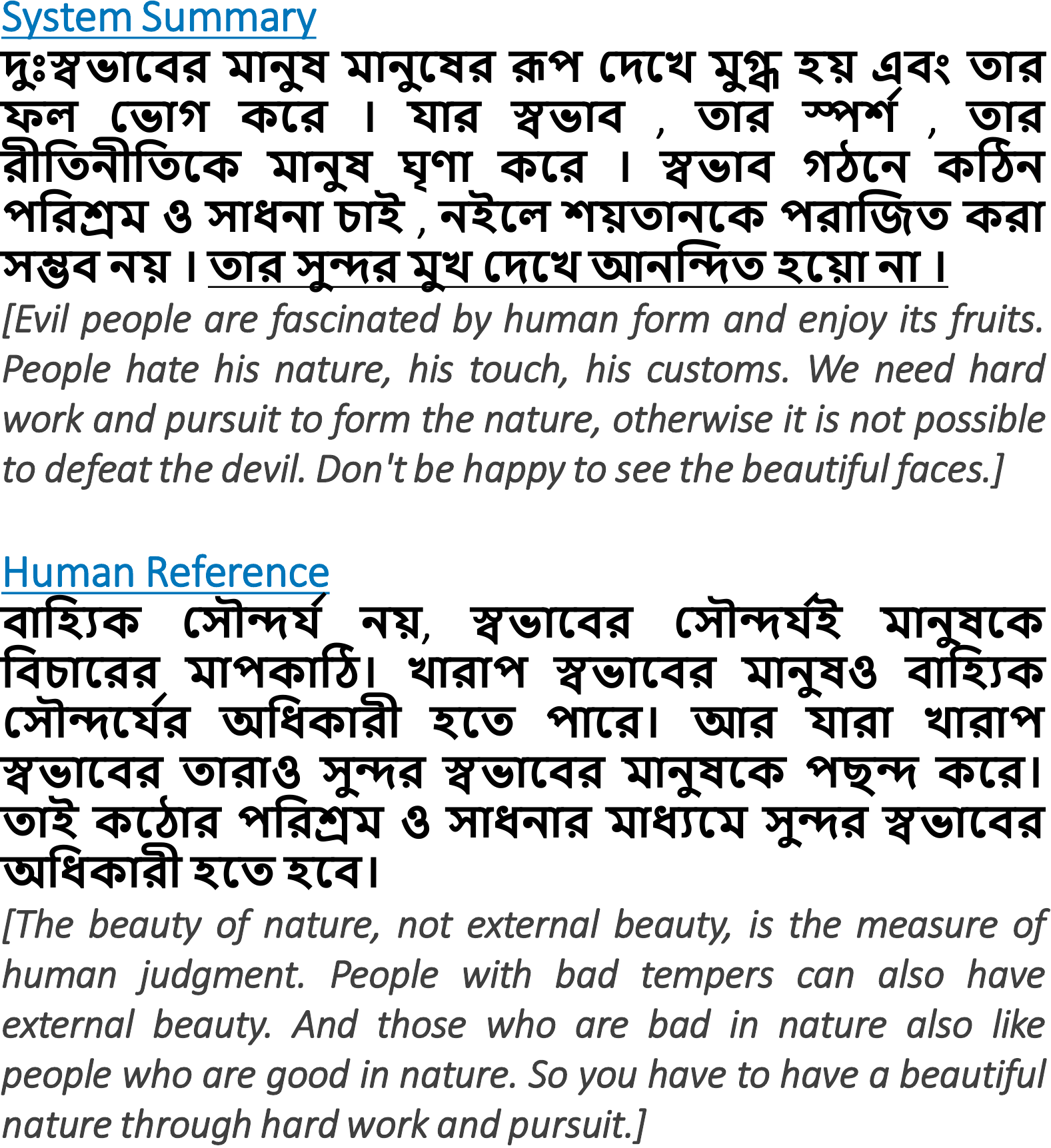}
    \caption{Example output of our \textbf{BenSumm} model with English translations.}
    \label{fig:sample_output}
\end{figure}

\section{Experiments}

This section presents our experimental details for assessing the performance of the proposed \textbf{BenSumm} model.

\paragraph{Dataset}

We conduct experiments on our dataset which consists of \emph{139} samples of human-written abstractive document-summary pairs written by professional summary writers of the National Curriculum and Textbook Board (\textbf{NCTB}). The NCTB is responsible for the development of the curriculum and distribution of textbooks. The majority of Bangladeshi schools follow these books.\footnote{\url{https://w.wiki/ZwJ}} We collected the human written document-summary pairs from the several printed copy of NCTB books. The overall statistics of the datasets are presented in Table \ref{tab:dataset-stats}. From the dataset, we measure the \emph{copy rate} between the source document and the human summaries. It's clearly visible from the table that our dataset is highly abstractive and will serve as a robust benchmark for this task's future works. Moreover, to provide our proposed framework's effectiveness, we also experiment with an extractive dataset \textbf{BNLPC}\footnote{\url{http://www.bnlpc.org/research.php}} \cite{haque2015automatic}. We remove the abstractive sentence fusion part to compare with the baselines for the extractive evaluation.

\paragraph{Automatic Evaluation}

We evaluate our system (\textbf{BenSumm}) using an automatic evaluation metric ROUGE F1 \cite{lin-2004-rouge} without any limit of words.\footnote{\url{https://git.io/JUhq6}} We extract $3$-best sentences from our system and the systems we compare as baselines. We report unigram and bigram overlap (ROUGE-1 and ROUGE-2) to measure informativeness and the longest common subsequence (ROUGE-L) to measure the summaries' fluency.  Since ROUGE computes scores based on the lexical overlap at the surface level, there is no difference in implementation for summary evaluation of the Bengali language.

\input{tables/datasets-stats}

\input{tables/results}

\begin{figure*}[htbp]
    \centering
    \includegraphics[scale = 0.30]{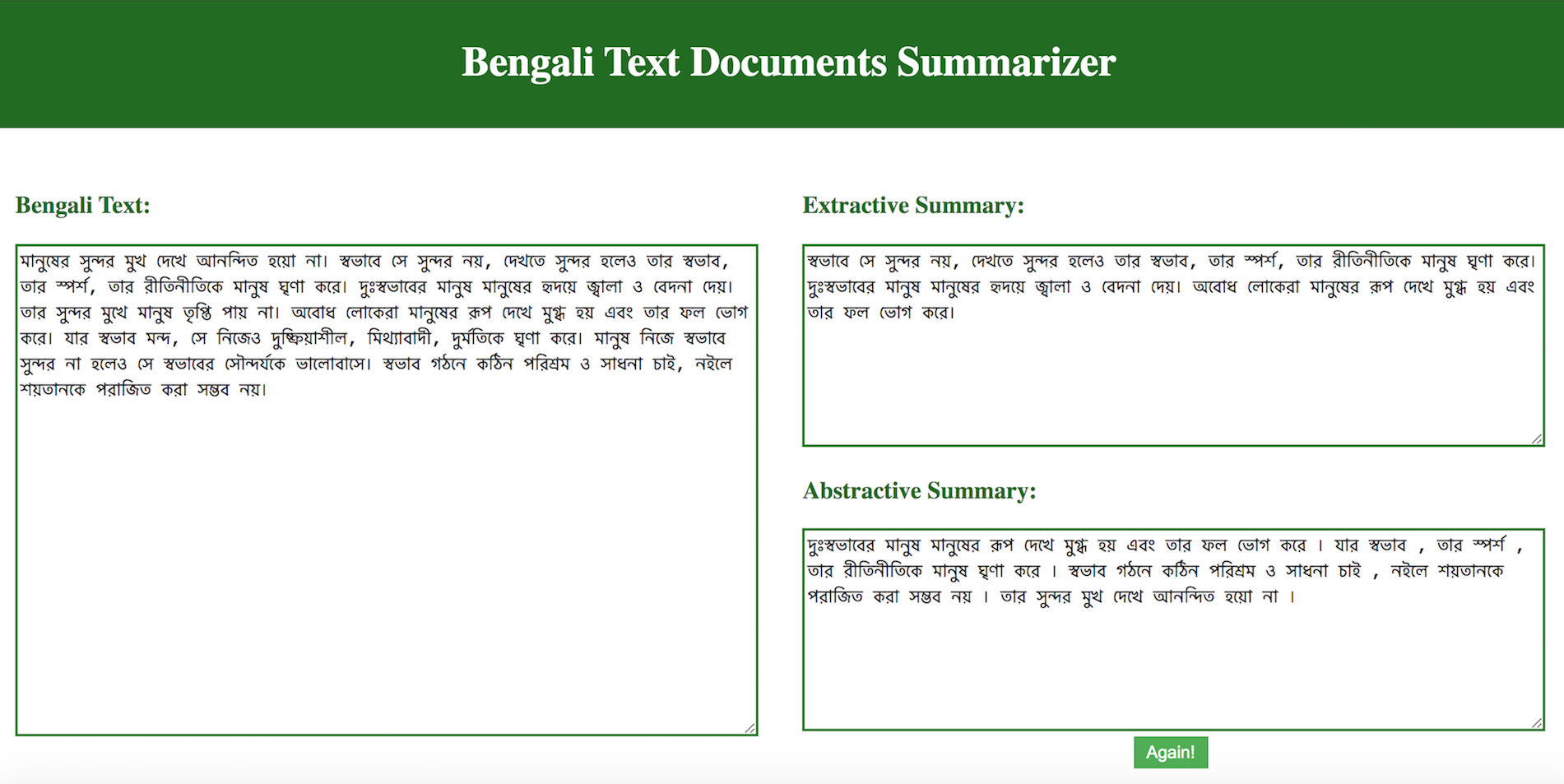}
    \caption{Interface of our Bengali Document Summarization tool. For an input document $D$ with $N$ sentences, our tool can provide both extractive and abstractive summary for the given document. The translations of both the document and summary are provided in the Appendix (see Figure \ref{fig:complete_process}).}
    \label{fig:tool}
\end{figure*}

\paragraph{Baseline Systems}
We compare our system with various well established baseline systems like \textbf{LexRank} \cite{erkan2004lexrank}, \textbf{TextRank} \cite{mihalcea2004textrank}, \textbf{GreedyKL} \cite{haghighi2009exploring}, and \textbf{SumBasic} \cite{nenkova2005impact}.  We use an open source implementation\footnote{\url{https://git.io/JUhq1}} of these summarizers and adapted it for Bengali language. It is important to note that these summarizers are completely extractive and designed for English language. On the other hand, our model is unsupervised and abstractive. 

\paragraph{Results}

We report our model’s performance compared with the baselines in terms of F1 scores of R-1, R-2, and R-L in Table \ref{tab:results}. According to Table \ref{tab:results}, our abstractive summarization model outperforms all the extractive baselines in terms of all the ROUGE metrics even though the dataset itself is highly abstractive (reference summary contains almost 73\% new words). Moreover, we compare our extractive version of our model \textbf{BenSumm} without the sentence fusion component. We get better scores in terms of R1 and RL compared to the baselines.  Finally, we present an example of our model output in Figure \ref{fig:sample_output}. Moreover, We design a Bengali Document Summarization tool (see Figure \ref{fig:tool}) capable of providing both extractive and abtractive summary for an input document.\footnote{Video demonstration of our tool can be accessed from \url{https://youtu.be/LrnskktiXcg}} 

\paragraph{Human Evaluation}

Though ROUGE \cite{lin-2004-rouge} has been shown to correlate well with human judgments, it is biased towards surface level lexical similarities, and this makes it inappropriate for the evaluation of abstractive summaries. Therefore, we assign three different evaluators to rate each summary generated from our abstractive system (\emph{BenSumm [Abs]}) considering three different aspects, i.e., Content, Readability, and Overall Quality. They have evaluated each system generated summary with scores ranges from 1 to 5, where 1 represents very poor performance, and 5 represents very good performance. Here, content means how well the summary can convey the original input document's meaning, and readability represents the grammatical correction and the overall summary sentence coherence. We get an average score of \textbf{4.41}, \textbf{3.95}, and \textbf{4.2} in content, readability, and overall quality respectively.

\section{Conclusion and Future Work}

In this paper, we have developed an unsupervised abstractive text summarization system for Bengali text documents. We have implemented a graph-based model to fuse multiple related sentences, requiring only a \textbf{POS} tagger and a pre-trained language model. Experimental results on our proposed dataset demonstrate the superiority of our approach against strong extractive baselines. We design a Bengali Document Summarization tool to provide both extractive and abstractive summary of a given document. One of the limitations of our model is that it cannot generate new words. In the future, we would like to jointly model multi-sentence compression and paraphrasing in our system. 

\section*{Acknowledgments}

We want to thank all the anonymous reviewers for their thoughtful comments and constructive suggestions for future improvements to this work.

\bibliography{eacl2021}
\bibliographystyle{acl_natbib}

\appendix

\section{Appendix}

\begin{figure*}[htbp]
    \centering
    \includegraphics[scale = 0.89]{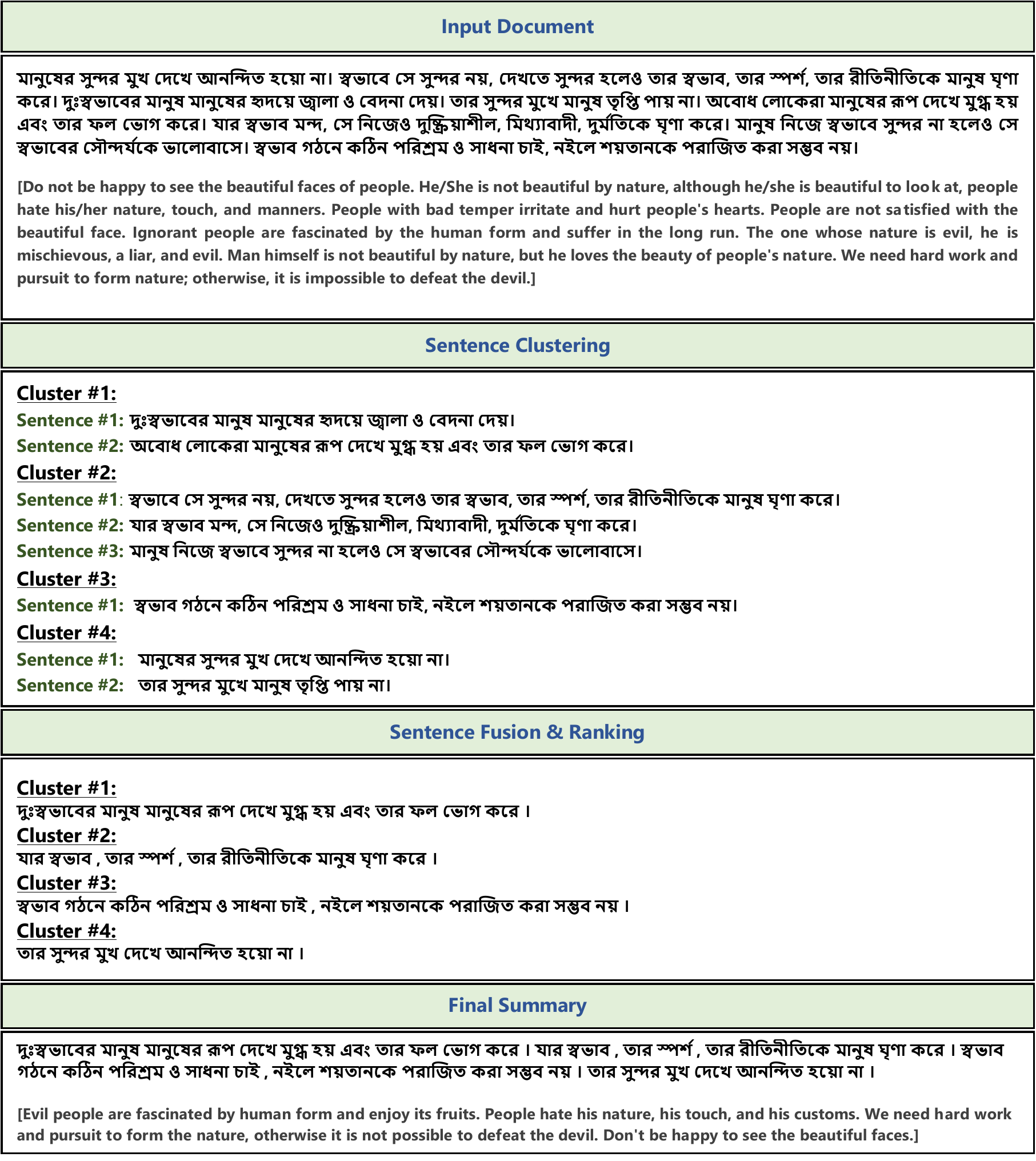}
    \caption{A detailed illustration with outputs from each step of our Bengali Abstractive Summarization model for a sample input document.}
    \label{fig:complete_process}
\end{figure*}

A detailed illustration of our \textbf{BenSumm} model with outputs from each step for a sample input document is presented in Figure \ref{fig:complete_process}.

\end{document}

%% file: tables/datasets-stats.tex
\begin{table}
\centering
\small
\begin{tabular}{@{}ccc@{}}
\toprule
\textbf{}                  & \textbf{\begin{tabular}[c]{@{}c@{}}NCTB\\ {[}Abstractive{]}\end{tabular}} & \textbf{\begin{tabular}[c]{@{}c@{}}BNLPC\\ {[}Extractive{]}\end{tabular}}  \\ \midrule \hline
Total \#Samples            & 139                                                                       & 200                                                                       \\
Source Document Length     & 91.33                                                                     & 150.75                                                                    \\
Human Reference Length     & 36.23                                                                     & 67.06                                                                     \\
\textbf{Summary Copy Rate} & \textbf{27\%}                                                             & \textbf{99\%}                                                             \\ \bottomrule
\end{tabular}
\caption{Statistics of the datasets used for our experiment. Length is expressed as Avg. \#tokens.}
\label{tab:dataset-stats}
\end{table}

%% file: tables/results.tex
\begin{table}
\centering
\begin{tabular}{@{}cccc@{}}
\toprule
\textbf{NCTB {[}Abstractive{]}} & \textbf{R-1}   & \textbf{R-2}   & \textbf{R-L}   \\ \midrule \hline
Random Baseline                 & 9.43           & 1.45           & 9.08           \\
GreedyKL                       & 10.01          & 1.84           & 9.46           \\
LexRank  & 10.65          & 1.78           & 10.04          \\
TextRank                        & 10.69          & 1.62           & 9.98           \\
SumBasic                       & 10.57          & 1.85           & 10.09          \\
BenSumm {[}Abs{]} (\textit{ours})              & \textbf{12.17} & \textbf{1.92}  & \textbf{11.35} \\ \midrule
\textbf{BNLPC {[}Extractive{]}} & \textbf{R-1}   & \textbf{R-2}   & \textbf{R-L}   \\ \midrule  \hline
Random Baseline                 & 35.57          & 28.56          & 35.04          \\
GreedyKL                       & 48.85          & 43.80          & 48.55          \\
LexRank                        & 45.73          & 39.37          & 45.17          \\
TextRank                        & 60.81          & \textbf{56.46} & 60.58          \\
SumBasic                       & 35.51          & 26.58          & 34.72          \\
BenSumm {[}Ext{]} (\textit{ours})               & \textbf{61.62} & 55.97          & \textbf{61.09} \\ \bottomrule
\end{tabular}
\caption{Results on our \textbf{NCTB} Dataset and \textbf{BNLPC}.}
\label{tab:results}
\end{table}